\pdfoutput=1

\documentclass[11pt]{article}

\usepackage[preprint]{acl}

\usepackage{times}
\usepackage{latexsym}
\usepackage{amsmath}
\usepackage{float}
\usepackage{multirow}
\usepackage{longtable}
\usepackage{subcaption}
\usepackage{arydshln}
\usepackage{hyperref}
\usepackage{multicol}

\usepackage[T1]{fontenc}

\usepackage[utf8]{inputenc}

\usepackage{microtype}

\usepackage{inconsolata}

\usepackage{graphicx}
\usepackage{times}
\usepackage{helvet}
\usepackage{courier}
\usepackage{graphicx}
\usepackage{amsfonts}
\usepackage{booktabs}

\title{MALoRA: Mixture of Asymmetric Low-Rank Adaptation for Enhanced Multi-Task Learning}

\author{
 \textbf{Xujia Wang\textsuperscript{1}  }\ 
 \textbf{Haiyan Zhao\textsuperscript{1}\thanks{Corresponding author.}  }\ 
 \textbf{Shuo Wang\textsuperscript{1}  }\ 
 \textbf{Hanqing Wang\textsuperscript{2}  }\ 
 \textbf{Zhiyuan Liu\textsuperscript{1}  }
\\
\\
 \textsuperscript{1}Tsinghua University\ \ \ \\
 \textsuperscript{2}Shanghai University of Finance and Economics
\\
}

\begin{document}
\maketitle
\begin{abstract}
Parameter-Efficient Fine-Tuning (PEFT) methods like LoRA have significantly improved the adaptation of LLMs to downstream tasks in a resource-efficient manner. 
However, in multi-task scenarios, challenges such as training imbalance and the seesaw effect frequently emerge.
Mixture-of-LoRA (MoLoRA), which combines LoRA with sparse Mixture-of-Experts, mitigates some of these issues by promoting task-specific learning across experts. 
Despite this, MoLoRA remains inefficient in terms of training speed, parameter utilization, and overall multi-task performance. 
In this paper, we propose \textbf{M}ixture of \textbf{A}symmetric \textbf{Lo}w-\textbf{R}ank \textbf{A}daptaion (MALoRA), a flexible fine-tuning framework that leverages asymmetric optimization across LoRA experts. MALoRA reduces the number of trainable parameters by 30\% to 48\%, increases training speed by 1.2x, and matches the computational efficiency of single-task LoRA models. 
Additionally, MALoRA addresses overfitting issues commonly seen in high-rank configurations, enhancing performance stability. 
Extensive experiments across diverse multi-task learning scenarios demonstrate that MALoRA consistently outperforms all baseline methods in both inter-domain and intra-domain tasks.\looseness-1
\end{abstract}

\section{Introduction}

Large Language Models (LLMs), such as BLOOM \cite{le2023bloom}, LLaMA \cite{touvron2023llama2,touvron2023llama}, and Mixtral 8x7B \cite{jiang2023mistral,jiang2024mixtral} have demonstrated remarkable general capabilities. 
These models, pre-trained on large and diverse datasets, can be adapted to new tasks through fine-tuning, leading to state-of-the-art performance in downstream applications~\cite{chung2024scaling,wei2021finetuned}, which extend their applicability across varied domains, significantly broadening their scope of use.\looseness-1

Fine-tuning LLMs by updating all parameters is computationally expensive and resource-intensive. To address this, PEFT methods such as Adapter \cite{houlsby2019parameter}, LoRA \cite{hu2021lora} and DoRA \cite{liu2024dora} were proposed. These methods focus on fine-tuning smaller, localized modules, significantly reducing memory usage and communication overhead. Despite their success, these methods still face limitations in complex multi-task scenarios.\looseness-1

Recent studies have highlighted several limitations of LoRA, such as training imbalances, catastrophic forgetting, and poor generalization to unseen tasks \cite{liu2024moe,luo2024moelora,liu2024adamole}. 
To address these issues, Mixture-of-LoRA (MoLoRA) was introduced, combining LoRA with a sparse Mixture-of-Experts (MoE) architecture. 
MoLoRA mitigates the training imbalances of LoRA by distributing tasks across multiple experts, allowing each expert to perform specialized functions and reducing the risk of catastrophic forgetting. The MoE router dynamically assigns weight coefficients to each LoRA expert, improving both task specialization and overall generalization in multi-task learning \cite{zadouri2023pushing}.
\looseness-1

\begin{figure*}[htb]
    \centering
    \includegraphics[width=1.0\linewidth]{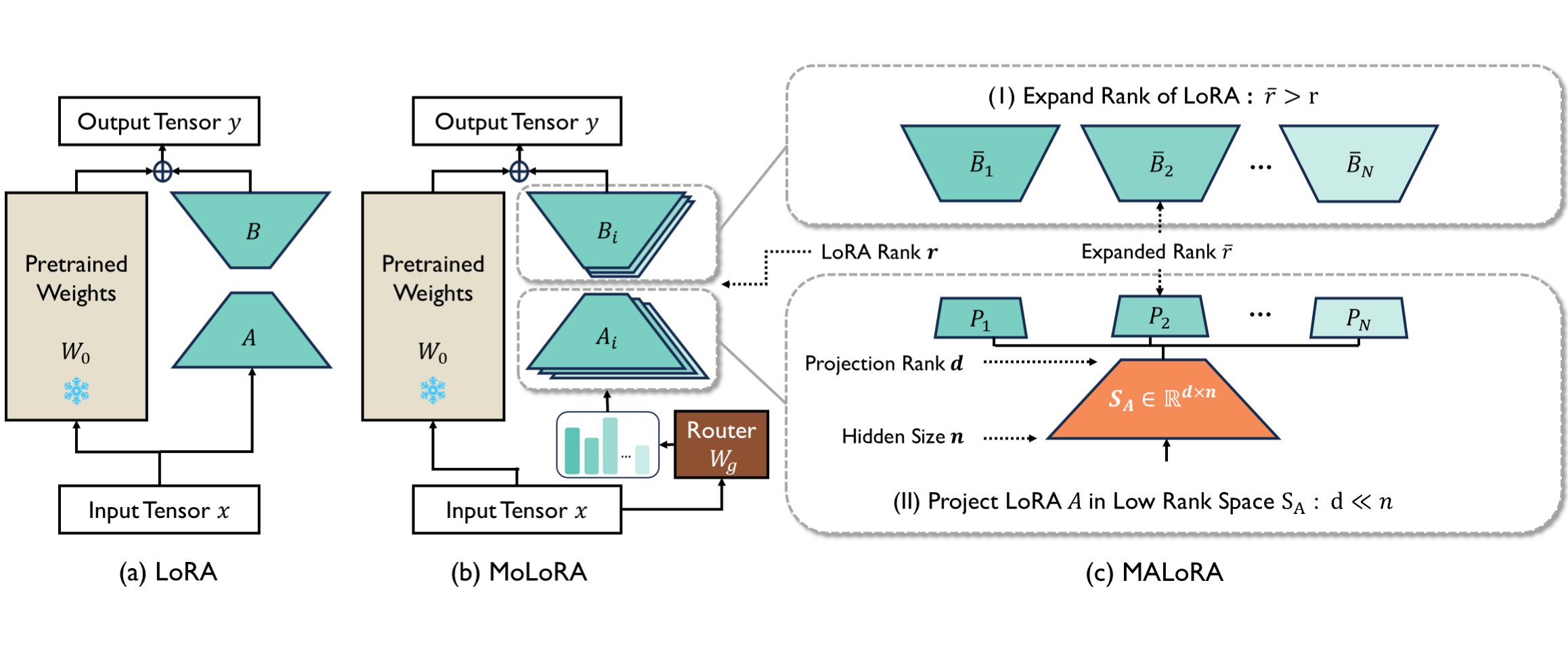}
    \vspace{-10mm}
    \caption{Architectures Overview: (a) LoRA, (b) MoLoRA, and (c) the proposed MALoRA. MALoRA optimizes the model in two key ways: (I) it increases the rank of the up-projection matrices ($B_t$) to enhance expert generalization capabilities, and (II) it introduces a shared low-rank subspace ($S_A$) in the down-projection matrices ($A_t$), while assigning each expert a unique coefficient matrix ($P_t$), effectively reducing parameter redundancy and computation.}
    \label{fig: MALoRA Structure}
    \vspace{-5mm}
\end{figure*}

However, MoLoRA introduces additional challenges, such as increased training latency and parameter redundancy. Existing works related to MoLoRA primarily focus on improving its routing strategies~\cite{dou2024loramoe,liu2024adamole,li2024mixlora} , yet they often overlook the intrinsic relationships and redundancies among LoRA experts. 
Inspired by the observed asymmetries in LoRA \cite{zhu2024asymmetry}, we conducted experiments to analyze the similarities and asymmetries between MoLoRA experts. 
Our findings reveal significant parameter redundancy, particularly in the down-projection matrices, where different experts show a high degree of similarity, indicating inefficiencies in its parameter usage. In contrast, the up-projection matrices exhibit much less similarity, suggesting the need for additional capacity to better capture task-specific variations.\looseness-1

To address these inefficiencies, we propose Mixture of Asymmetric Low-Rank Adaptation (MALoRA), as shown in Figure \ref{fig: MALoRA Structure}. MALoRA introduces a shared, tunable low-rank subspace in the down-projection module, with each LoRA expert assigned a compact coefficient matrix, effectively reducing parameter count and computational complexity while preserving distinctions between experts. By reallocating the parameters saved from the down-projection module to the up-projection module, MALoRA increases the rank of the up-projection matrices, enhancing the model’s theoretical generalization bounds. By leveraging these asymmetries, MALoRA optimizes parameter usage, reduces redundancy, and improves both generalization and multi-task learning performance.\looseness-1

In summary, our contributions are as follows:

(1) We identify and quantify the parameter redundancy in MoLoRA, particularly in the down-projection matrices, where experts demonstrate significant overlap. This insight highlights inefficiencies in the model’s current parameter allocation strategies, with distinct optimization needs identified for the up- and down-projection modules.\looseness-1

(2) We propose MALoRA, a novel fine-tuning framework that introduces a shared low-rank subspace in the down-projection module, reducing redundancy and computational overhead. The reallocated parameters are utilized to enhance the up-projection module, expanding its rank and improving the model’s theoretical generalization capacity.\looseness-1

(3) MALoRA reduces the number of trainable parameters by 30\% to 48\%, while maintaining or surpassing the performance of MoLoRA across multi-tasks. It further achieves a 1.2x speedup in training and inference, alleviating high-rank overfitting and optimizing computational efficiency.\looseness-1

(4) We conduct extensive evaluations across both inter-domain and intra-domain multi-task learning scenarios, demonstrating that MALoRA consistently outperforms all baseline methods in both efficiency and generalization, offering a scalable and robust solution for fine-tuning large models.\looseness-1

\section{Preliminaries}
\paragraph{LoRA}
Low-Rank Adaption \cite{hu2021lora} assumes that the updates to the linear weight $W\in \mathbb{R}^{m\times n}$ exhibit a low-rank structure. It employs two trainable low-rank matrices $A\in \mathbb{R}^{r\times n}$ and $B\in \mathbb{R}^{m\times r}$ to approximate the parameter update of $W$ during fine-tuning:\looseness-1

\begin{equation}
\Delta W = \frac{\alpha}{r} BA
\end{equation}

Here, $r$ represents the rank of decomposed matrices and $\alpha$ controls the scale of the update. Given that $r\ll \min\{n,m\}$, LoRA greatly reduce the memory and GPU communication overhead.\looseness-1

\paragraph{AsyLoRA}\label{sec:asymmetry lora}
Asymmetry LoRA \cite{zhu2024asymmetry} builds on LoRA by leveraging the inherent asymmetry between the low-rank matrices $A$ and $B$. Specifically, $A$ tends to extract features from the input while $B$ refines these features to align with task-specific objectives.\looseness-1

AsyLoRA enhances LoRA by focusing the computational budget on the up-projection matrix $B$, keeping it fully trainable while freezing the parameters of matrix $A$. This method effectively doubles the rank $r$ without increasing the overall trainable parameter count. The work also brought forward a generalization boundary for LoRA. Given a distribution $\mu$ and a trainable parameter set $\mathcal{A}$, is reformulated as follows. Here, $\gamma$ is a constant influenced by training hyperparameters, and $|\mathcal{A}|$ represents the number of trainable parameters:\looseness-1
\begin{equation}
|gen(\mu, \mathcal{A})| \leq \sqrt{\gamma |\mathcal{A}|}
\end{equation}

\paragraph{Mixture-of-LoRA}
Mixture-of-LoRA \cite{zadouri2023pushing} (MoLoRA) is a novel PEFT framework that integrates MoE architecture with LoRA modules serving as the experts. The built-in router of MoE dynamically distributes input data to different LoRA experts. 
Specifically, a MoLoRA layer consists of $N$ independent LoRA experts $\{E_t\}_{t=1}^{N}$. The router, parameterized by a learnable weight matrix $W_g$, assigns routing weights $G_t$ to each expert $t$ based on the input tensor $x$. The routing decision and the the layer output are computed as follows:\looseness-1

\vspace{-5mm}
\begin{align}
&\Delta W_t = B_t A_t \\
&G_t = \text{TopK}(\text{Softmax}(W_g\cdot x)) \\
&y = \sum_{i=t}^N G_t\Delta W_t x
\end{align}

where $G_t$ represents the top $K$ experts selected by the router, ensuring that only the most relevant experts are activated. 
This dynamic allocation allows MoLoRA to efficiently handle multi-task scenarios by assigning appropriate experts to each task. 
MoLoRA has demonstrated strong generalization to unseen tasks and excels in multi-task learning settings~\cite{luo2024moelora,li2024mixlora}.\looseness-1

\section{Method}

We propose MALoRA, which builds on the observed asymmetries in LoRA experts within the MoLoRA architecture. MALoRA improves parameter efficiency and enhances generalization in multi-task learning by leveraging both the similarities and asymmetries in the down-projection and up-projection matrices. This approach reduces the number of trainable parameters by at least 30\% and reallocates the saved resources to expand the model’s generalization capacity, all without adding significant computational overhead.\looseness-1

\subsection{Asymmetry in MoLoRA}\label{sec:asymmetry molora}

\begin{figure}[t]
    \centering
    \includegraphics[width=1.0\linewidth]{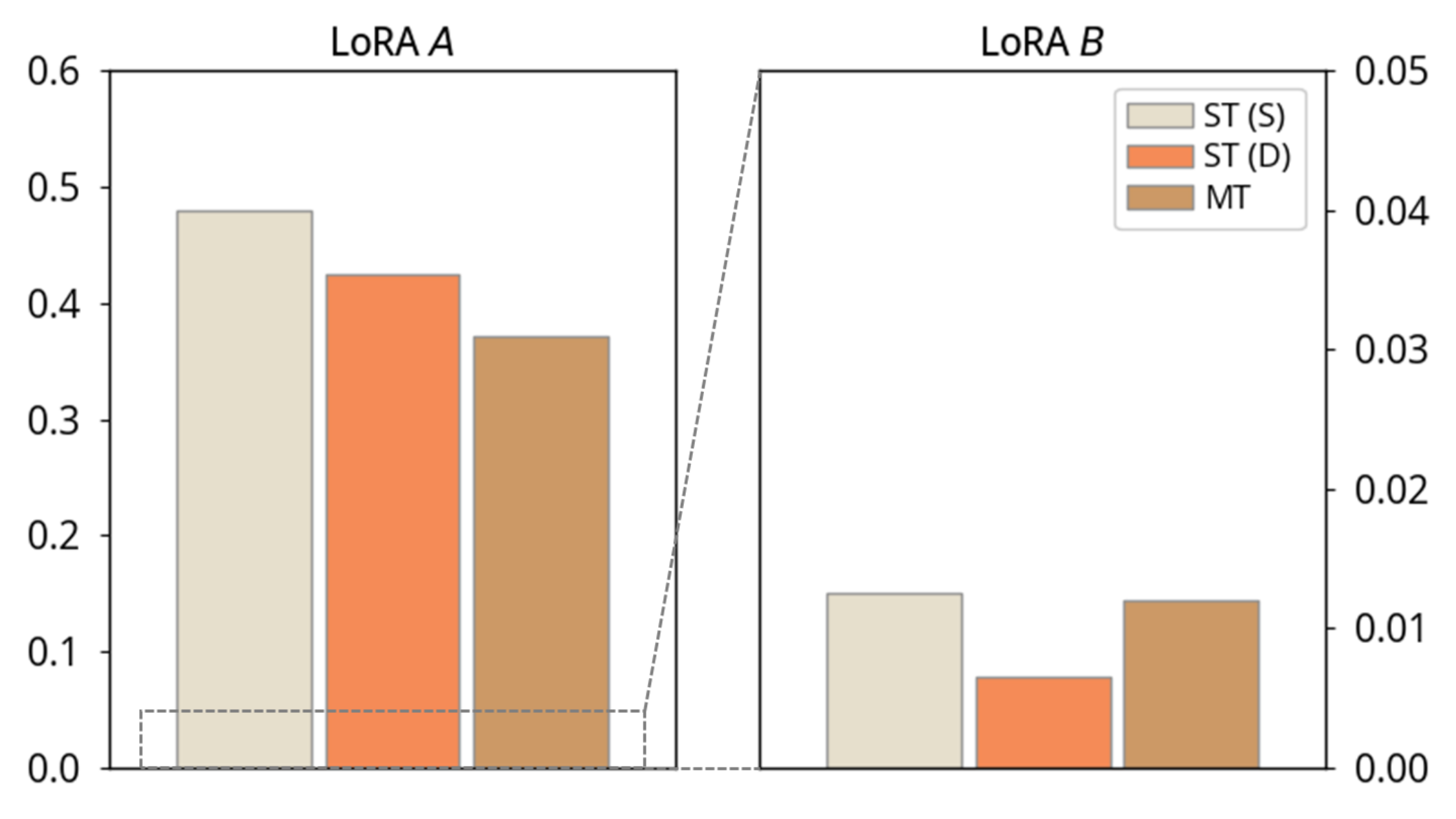}
    \caption{Spatial Similarity Analysis. Spatial similarity between LoRA experts within the same MoLoRA layer, evaluated using CCA. The down-projection matrix ($A$) demonstrates significantly higher similarity across all learning scenarios (ST(S), ST(D), MT), suggesting it captures generalized features. In contrast, the up-projection matrix ($B$) shows much lower similarity, indicating its role in task-specific fine-tuning.}
    \label{fig: Spatial Similarity}
    \vspace{-15pt}
\end{figure}

To explore the inter-expert relationships in MoLoRA, we analyzed the spatial similarity between the down-projection ($A$) and up-projection ($B$) matrices across different tasks. As illustrated in Figure \ref{fig: Spatial Similarity}, the down-projection matrices ($A$) exhibit significantly higher similarity compared to the up-projection matrices ($B$), indicating a notable asymmetry in their behavior.\looseness-1

We performed a Canonical Correlation Analysis (CCA) \cite{ramsay1984matrix} to evaluate the similarities between LoRA experts under three different learning scenarios: same task (ST(S)), different tasks (ST(D)), and multi-task learning (MT). The results show that $A$ consistently maintains higher similarity across all scenarios. 
As task complexity increases from ST(S) to MT, the similarity of $A$ decreases, reflecting its adaptability to more diverse task settings. Despite this decline, $A$ still retains relatively high similarity, indicating that it captures generalized patterns while adapting to task-specific nuances by marginal differences.
In contrast, $B$ demonstrates much lower similarity, with values below 0.05 across all scenarios, suggesting that it is more focused on task-specific fine-tuning rather than generalization.\looseness-1

\begin{figure}[t]
    \centering
    \includegraphics[width=1.0\linewidth]{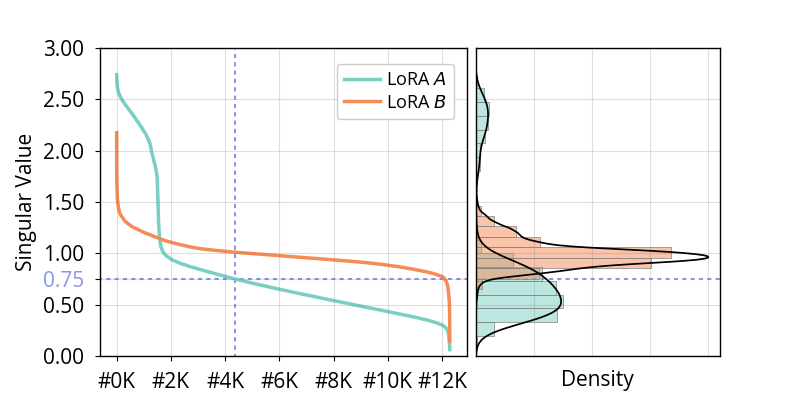}
    \caption{Singular Values of the Concatenated Homologous Matrices in descending order. matrix $B$ shows a concentration of larger singular values, indicating that many singular vectors are important for task-specific fine-tuning. In contrast, $A$ has more smaller singular values, with only a few larger ones, suggesting that only a subset of singular vectors play a critical role. This reflects that $B$ distributes importance across more components, while $A$ relies on a smaller, more focused set of key features for generalization.\looseness-1}
    \label{fig: Singular Value}
    \vspace{-13pt}
\end{figure}

The singular value distribution in Figure \ref{fig: Singular Value} illustrates the different roles of the down-projection ($A$) and up-projection ($B$) matrices. For $B$, the singular values are closely clustered within a narrow range between 0.75 and 1.25, indicating that most basis vectors contribute similarly and are uniformly important for task-specific fine-tuning. In contrast, $A$ exhibits a broader range of singular values, spanning from 0.25 to 2.75. This wider distribution suggests that a smaller subset of basis vectors with larger singular values (approximately 33\% of the total) play a dominant role in capturing key features. The remaining vectors, associated with smaller singular values, contribute less significantly. This indicates that $A$ can be effectively optimized by focusing on the vectors with larger singular values, allowing the model to capture both generalizable and task-specific information more efficiently.\looseness-1

\subsection{Projecting $A$ in Low-Rank Space}

Given the observed similarities among the down-projection matrices $A_t$ across different LoRA experts, MALoRA optimizes $A_t$ by projecting it into a layer-shared low-rank subspace. This subspace is defined by a matrix $S_A \in \mathbb{R}^{d \times n}$, which is shared across all experts in a MoLoRA layer. This allows for more efficient feature extraction while reducing redundancy among experts.\looseness-1

Instead of treating each expert’s down-projection matrix independently, MALoRA projects each $A_t$ onto the span of $S_A$, capturing the most relevant features in a lower-dimensional space. It is verified in the previous section, that the model only needs to learn the basis vectors with the larger singular values, which constitute 33\% of the total. Thereby the rank $d$ of $S_A$ is kept small relative to the original matrix dimensions, enabling efficient compression of $A_t$ without significant loss in expressiveness. Each expert’s matrix $A_t$ is then replaced by a projection coefficient matrix $P_t \in \mathbb{R}^{r \times d}$, which adjusts the contributions of the shared $S_A$ basis vectors for each expert to better align with task-specific needs.\looseness-1

Mathematically, for a MoLoRA layer with $N$ experts, MALoRA redefines the parameter set as $\{W_g, S_A\} \cup \{(P_t, \overline{B}_t)\}_{t=1}^N$, where $\overline{B}_t$ is the up-projection matrix (details provided in the next sub-section). 
During forward propagation, the down-projection matrix $A_t$ is replaced by its projection onto $S_A$, leading to the following expressions:\looseness-1
\begin{align}
&\Pi_{\text{span}(\vec{S_A})}(A_t) = P_t S_A\\
&\Delta W_t = \overline{B}_t P_t S_A
\end{align}

For initialization, the up-projection matrix $\overline{B}_t$ is set to zero, consistent with the LoRA approach. While the down-projection matrices in LoRA are initialized using the Kaiming Uniform initialization \cite{he2015delving} to ensure proper gradient scaling, we aim for the product $P_t S_A$ to similarly follow a Kaiming Uniform distribution as well. To achieve this, we apply singular value decomposition (SVD) to several Kaiming Uniform-initialized random matrices $K_t \in \mathbb{R}^{d \times n}, t \in [0, N]$, where each expert receives its own initialized matrix. Since $K_t$ has a rank $d$ ($d<n$), the singular vectors beyond rank $d$ are zero-filled and cropped from the SVD results, as shown in Eq.~\ref{eq:pt1}.\looseness-1

The product of \( U_t \) and \( \Sigma_t \) is clipped to rank \( r \) for initialization of $P_t$, $S_A$ is initialized using the right singular matrix $V_0$ from the first SVD decomposition ($A_0$), as it serves as a common basis across all experts within the layer.
This initialization establishes a global subspace shared by all experts, while  $P_t$  is responsible for capturing task-specific adaptations. 
The initialization is given by:\looseness-1
\begin{align}
(U_t)_{d\times d}, (\Sigma_t)_{d\times d}, (V_t)_{d\times n} &= \text{SVD}_{\text{crop}}(K_t)\label{eq:pt1}\\
\centering
P_t = (U_t \Sigma_t)&\text{[:r]} / {\beta}\\
S_A = \beta V_0&
\end{align}
where $\beta$ is a hyperparameter that controls the balance between general and task-specific learning. A higher $\beta$ amplifies the gradients of $P_t$, thereby enhancing the model's ability to focus on task-specific details. Conversely, a lower $\beta$ shifts the emphasis toward capturing generalized features by giving more weight to $S_A$.
Please refer to Appendix \ref{sec:Impact of beta} and \ref{sec:Rank Variation Ablation Study of MALoRA} for detailed analysis.\looseness-1

\subsection{Expanding Rank of LoRA Experts}
\label{sec: expand}

In Section \ref{sec:asymmetry molora}, we discussed reducing the dimensionality of the down-projection matrices $A_t$. By setting the rank of the shared matrix $S_A$ to $d = \lambda rN$, where $\lambda < 1$. $r$ is the rank of each LoRA expert, and $N$ is the number of experts, we eliminate the need for separate down-projection matrices. This reduction reallocates resources to expand the up-projection matrix $B$, increasing its rank to $\overline{r} = r + (1 - \lambda)r$ for each LoRA expert.\looseness-1

The computational savings from compressing $A_t$ are reallocated to expand the up-projection matrices $\overline{B}_t$. The newly introduced parameter matrix $P_t$, sized at only $\overline{r} \times d$, is relatively small compared to the model’s hidden size $n$, contributing less than 1\% to the total parameter count of PEFT. Thus, the additional parameters introduced by MALoRA are negligible.
By expanding the rank of LoRA experts, MALoRA enhances the model's generalization capability. According to the generalization bound derived in Section \ref{sec:asymmetry lora}, the upper bound $\mathcal{U}$ is given by:\looseness-1
\begin{align}\label{eq:bound}
\mathcal{U}(|gen(\mu,  \mathcal{A})|) \propto \sqrt{|\mathcal{A}|} \propto \sqrt{r}
\end{align}

For each expert, as the ranks of $S_A$, $P_t$, and $\overline{B}_t$ all exceed $\overline{r}$, the resulting rank of $\Delta W_t$ is $\overline{r}$, whereas MoLoRA experts retain the original rank $r$. MALoRA establishes a generalization boundary that is $\sqrt{\overline{r}/r}$ times greater than that of MoLoRA, resulting in improved performance across tasks.\looseness-1

\section{Experiments}

In this section, we evaluate MALoRA’s performance across diverse multi-task learning scenarios. We describe the datasets used for training and evaluation, outline the implementation details, and compare MALoRA with baseline methods. The results highlight MALoRA’s efficiency and accuracy improvements, followed by an ablation study to examine the impact of key components.\looseness-1

\subsection{Experimental Setup}

\paragraph{Datasets}

To evaluate MALoRA, we conducted experiments in both inter-domain and intra-domain settings.
We selected datasets from different domains, including MetaMathQA, Magicoder, MedMCQA, and Finance Alpaca, representing tasks such as math reasoning, code generation, medical knowledge, and finance, respectively. Additionally, we included the E2E dataset for specialized task evaluation and Alpaca-GPT4 to test common-sense reasoning and instruction-following. The multi-domain training set was created by sampling 30,000 instances from each dataset and blending them uniformly. For evaluation, we used GSM8K for math, HumanEval for code, Financial PhraseBank for finance, and ARC for common-sense reasoning, along with task-specific test splits for other domains.\looseness-1

For intra-domain evaluation, we focused on multi-task learning within the domain of common-sense reasoning. We employed datasets like PIQA, OBQA, BoolQ, and ARC to evaluate MALoRA’s performance on closely related tasks. This allows us to measure the model’s generalization within a specific domain, assessing how effectively it handles tasks with similar data distributions.
Details of datasets can be found in Appendix \ref{sec:Datasets}.\looseness-1

\begin{table*}[ht]
\centering
\renewcommand{\arraystretch}{1.3}
\resizebox{\textwidth}{!}{
\begin{tabular}{lcccccccccc}
\specialrule{2pt}{0pt}{0pt}
Method                       & \#Params & Latency($\mu s$) & MedMCQA & HumanEval & GSM8K & PhraseBank & ARC-C & ARC-E & E2E & AVG. \\ \hline
LLaMA-2 7B     & - & - & 34.2 & 14.6 & 13.2 & 57.6 & 43.1 & 75.5 & 21.7 & 37.1       \\ \hline
LoRA (ST)        & 2.1\% & \multirow{2}{*}{-} & 42.5 & 34.8 & 36.5 & 77.4 & 62.4 & 79.3 & 66.4 & 57.0             \\
AsyLoRA (ST)     & 2.2\% & & 42.6 & 30.5 & 33.1 & 78.8 & 61.2 & 77.4 & 66.0 & 55.7           \\ \hline
LoRA             & \underline{2.1\%} & \textbf{833.3} & \underline{43.2} & \textbf{34.8} & \underline{34.0} & 69.6 & 59.1 & 76.5 & 65.9 & 54.7      \\
AsyLoRA         & 2.2\% & - & \textbf{43.3} & 31.1 & 27.6 & 74.2 & 59.8 & \underline{77.8} & 65.5 & 54.2      \\
DoRA             & \underline{2.1\%} & 1020.9 & 41.3 & 32.3 & 32.6 & 74.0 & \underline{60.0} & 75.6 & \textbf{66.5} & 54.6      \\ \hdashline[2pt/4pt]
MoAsyLoRA        & 2.3\% & - & 40.6 & 24.4 & 23.9 & 73.0 & 58.5 & 76.0 & 65.8 & 51.7      \\
MoLoRA           & 2.2\% & 1072.3 & 42.1 & 29.3 & 33.1 & 78.1 & \textbf{61.1} & 78.1 & 65.6 & 55.3      \\ \hdashline[2pt/4pt]
\textbf{MALoRA-Small}      & \textbf{1.6\%} & \multirow{2}{*}{\underline{896.4}} & 41.0 & 29.9 & \underline{34.0} & \textbf{80.3} & 59.4 & 77.2 & 65.7 & \underline{55.4}      \\ 
\textbf{MALoRA}      & 2.3\% &  & 42.3 & \underline{32.9} & \textbf{34.1} & \underline{79.8} & \underline{60.0} & \textbf{78.5} & \underline{66.3} & \textbf{56.3}      \\ 
 \specialrule{2pt}{0pt}{0pt}
\end{tabular}
}
\caption{Comparison of various PEFT methods for multi-task learning across different domains. "ST" stands for single downstream task fine-tuning. For the E2E task, the ROUGE-L metric is used, while accuracy is reported for others. "\#Params" refers to the percentage of trainable parameters relative to the base model. MALoRA and MALoRA-Small achieve the best results with the lowest latency, highlighting their efficiency.\looseness-1}
\label{tag: cross domain result}
\vspace{-10pt}
\end{table*}

\paragraph{Implementation Details}

We adopt LLaMA-2 7B as the backbone model and compare MALoRA against LoRA, DoRA, Asymmetry LoRA, and MoLoRA. For both LoRA and DoRA, the rank $r$ is set to 64, with a dropout rate of 0.05. Asymmetry LoRA doubles the rank, keeping matrix $A$ frozen during fine-tuning. Following mainstream practices, MoLoRA uses 8 LoRA experts, each with a rank of $r=8$, and activates the top-2 experts based on input routing, with an auxiliary loss factor of 0.001. MALoRA expands the rank to $\overline{r}=12$, and the shared matrix $S_A$ has a rank of $d=32$, ensuring parameter comparability across all baselines. Additionally, MALoRA-Small is introduced, where $\overline{r}=8$ and $d=22$, providing a more parameter-efficient variant for comparison.
The PEFT modules are applied to all linear layers within the transformer architecture. All backbone parameters remain frozen throughout the experiments. For inter-domain learning tasks, the learning rate is set to $5e^{-4}$ with a batch size of 4, while intra-domain tasks, the learning rate is $4e^{-4}$ with a batch size of 2. These hyperparameters were determined via grid search to optimize performance across tasks.
Further details can be found in Appendix \ref{sec:Hyper-parameter Configuration}.\looseness-1

\subsubsection{Main Result}

\paragraph{Inter-domain Performance}
As shown in Table~\ref{tag: cross domain result}, MALoRA outperforms all baselines in inter-domain settings, achieving an average score of $56.3$, surpassing both LoRA and MoLoRA by 1.6\% and 1.0\% respectively. This performance is particularly notable on tasks like GSM8K and HumanEval, where MALoRA demonstrates its capacity to handle task complexity. These gains are attributed to MALoRA’s effective integration of an asymmetric low-rank adaptation and multi-expert MoE structure, which extends the generalization boundary of the model while maintaining a low computational overhead.
Baselines without an MoE structure struggled on tasks like Finance PhraseBank due to dataset imbalance, where label-heavy tasks like Magicoder dominated, leading to underfitting in minority datasets such as financial data. This highlights the seesaw phenomenon often seen in traditional PEFT methods. In contrast, MALoRA’s ability to handle diverse multi-domain tasks showcases its robustness in orthogonal domain learning with significant distribution differences. 
MALoRA-Small, with only 1.6\% trainable parameters, uses fewer parameters than any other method while achieving an impressive average performance. Although its overall results are slightly lower than MALoRA, MALoRA-Small still outperforms all other multi-task methods. This result emphasizes MALoRA’s ability to scale down without sacrificing much performance, making it an ideal solution for scenarios where computational efficiency is a priority. More analysis are available in Appendix \ref{sec:Quantity of Training Tokens} and \ref{sec:Insufficient Learning Ability of AsyLoRA}. \looseness-1

\begin{table}[b]
\vspace{-5mm}
\centering
\renewcommand{\arraystretch}{1.4}
\resizebox{\columnwidth}{!}{%
\begin{tabular}{lcccccc}
 \specialrule{1pt}{0pt}{0pt}
Method              & PIQA & OBQA & BoolQ  & ARC-C & ARC-E & AVG. \\ \hline
        LoRA & \underline{81.1} & 89.4 & 77.1 & 64.5 & 80.5 & 78.48 \\
        AsyLoRA & 80.4 & 87.4 & \textbf{82.0} & 64.9 & \underline{82.4} & 79.42 \\ 
        DoRA & 80.0 & 89.2 & 80.4 & 65.9 & 82.4 & 79.56 \\ 
        MoLoRA & 80.3 & \textbf{90.4} & 79.3 & \underline{66.1} & \underline{82.4} & \underline{79.69} \\ 
        \textbf{MALoRA} & \textbf{81.8} & \underline{89.6} & \underline{81.6} & \textbf{66.6} & \textbf{82.8} & \textbf{80.47} \\
 \specialrule{1pt}{0pt}{0pt}
\end{tabular}%
}
\caption{Comparison of different PEFT methods on intra-domain common-sense reasoning tasks. MALoRA achieves the highest average score, highlighting its effectiveness in handling tasks within the same domain.\looseness-1}

\label{tag: common sense result}
\end{table}

Additionally, the latency results clearly demonstrate MALoRA’s efficiency. It reduces training time by 16\% compared to MoLoRA, while achieving higher performance across most tasks. Notably, MALoRA’s latency is comparable to LoRA’s, despite incorporating a more complex multi-expert structure. This shows that MALoRA can offer significant performance improvements without incurring a substantial increase in computational cost.\looseness-1

\paragraph{Intra-domain Performance}
For common-sense reasoning tasks, as shown in Table \ref{tag: common sense result}, MALoRA again demonstrates its superiority. It improves upon LoRA by 2\% and MoLoRA by 0.8\%, proving its efficacy even in more focused intra-domain tasks. Notably, MALoRA’s robust performance across various tasks indicates that it strikes a good balance between learning domain-specific features and maintaining generalization capability. It also highlights MALoRA’s ability to manage data imbalance, as seen in tasks like ARC-C, where traditional methods like LoRA struggle.\looseness-1

\begin{table*}[h]
\centering
\renewcommand{\arraystretch}{1.0}
\resizebox{\textwidth}{!}{%
\begin{tabular}{lcccccccccc}
 \specialrule{2pt}{0pt}{0pt}
Method                        & \#Params & MedMCQA & HumanEval & GSM8K & PhraseBank & ARC-C & ARC-E & E2E & AVG. \\ \hline
        MoLoRA           & 2.2\% & 42.1 & 29.3 & 33.1 & 78.1 & 61.1 & 78.1 & 65.6 & 55.3 \\ 
        MoLoRA, $r=12$ & 3.3\% & 40.8 & 31.7 & \textbf{34.9} & 77.3 & \textbf{62.6} & 78.2 & 66.2 & 56.0 \\ 
        w same $A_t$, $r=12$ & 3.3\% & 42.4 & 29.9 & 33.7 & 79.1 & 61.9 & \textbf{78.8} & 66.3 & 56.0 \\ \hline
        \textbf{MALoRA} & 2.3\% & 42.3 & \textbf{32.9} & 34.1 & \textbf{79.8} & 60.0 & 78.5 & \textbf{66.3} & \textbf{56.3} \\ 
        w fixed $S_A$, $r=16$ & 2.4\% & 43.2 & 28.7 & 32.7 & 73.9 & 58.6 & 76.8 & 66.0 & 54.3 \\
        w/o Asymmetry & 2.2\% & 42.4 & 29.3 & 35.4 & \textbf{79.8} & 59.6 & 77.8 & 65.9 & 55.8 \\
        w/o shared $S_A$ & 2.3\% & \textbf{43.9} & 28.1 & 29.7 & 66.2 & 61.1 & 78.1 & 65.8 & 53.3 \\
        w/o ft $P_t$  & 2.3\% & 41.9 & 29.9 & 34.1 & 78.9 & 58.7 & 78.8 & 65.8 & 55.5 \\
        decomposing $B_t$ & 2.2\% & 43.0 & 30.5 & 32.1 & 74.1 & 61.3 & 76.6 & 65.9 & 54.8 \\
 \specialrule{2pt}{0pt}{0pt}
\end{tabular}%
}
\vspace{-5pt}
\caption{Ablation study results for MALoRA, showing the impact of different components on model performance. \looseness-1 }

\label{tag: ablation study}
\vspace{-5pt}
\end{table*}

\subsection{Ablation Study}

\begin{figure*}[htbp]
\centering
\begin{subfigure}{0.325\linewidth}
    \includegraphics[width=0.9\textwidth]{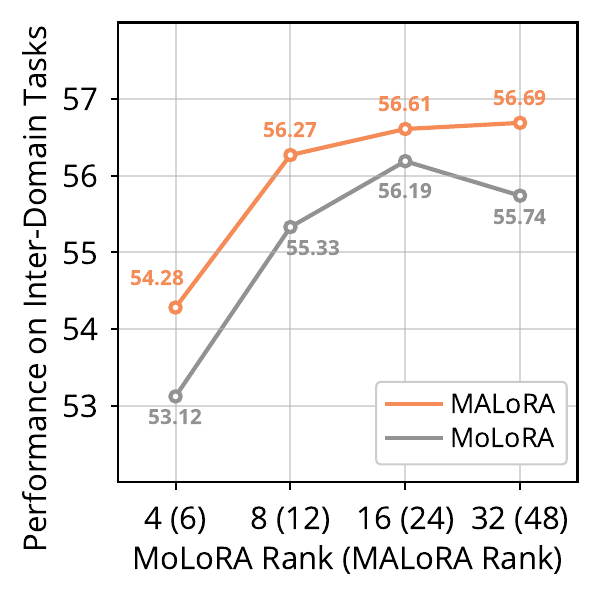}
    \label{fig:image1}
    \vspace{-5pt}
    \caption{Performance Across Ranks}
\end{subfigure}
\centering
\begin{subfigure}{0.325\linewidth}
    \includegraphics[width=0.9\textwidth]{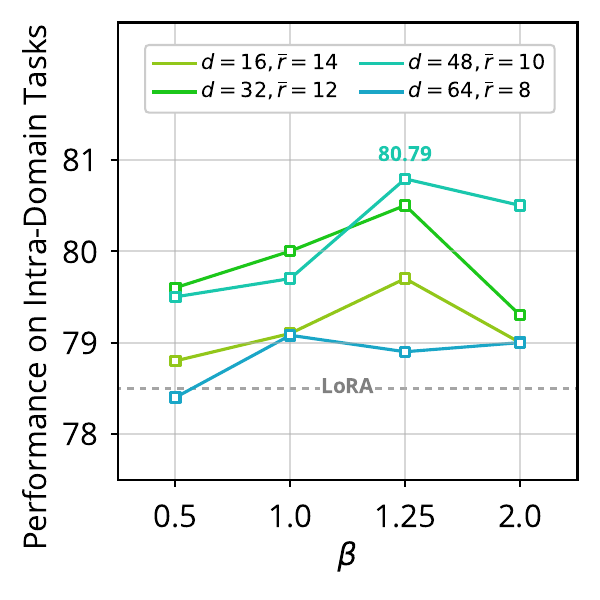}
    \label{fig:image2}
    \vspace{-5pt}
    \caption{Ablation Study of $\beta$ and $d$}
\end{subfigure}
\centering
\begin{subfigure}{0.325\linewidth}
    \includegraphics[width=0.9\textwidth]{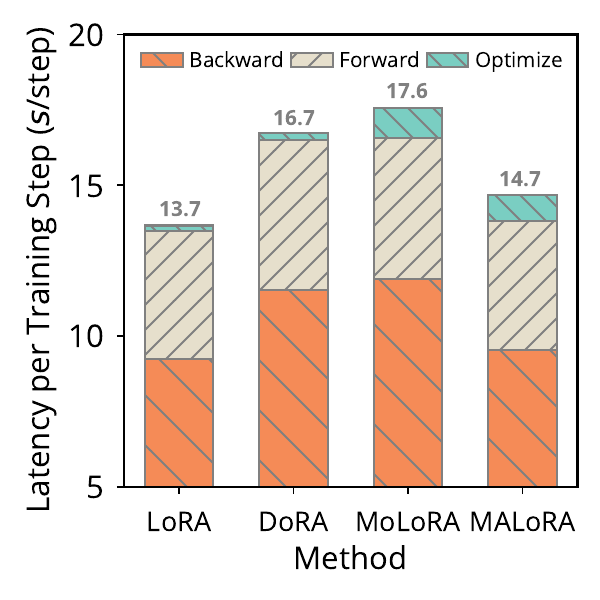}
    \label{fig:image3}
    \vspace{-5pt}
    \caption{Training Latency Comparison}
\end{subfigure}
\vspace{-5pt}
\caption{(a) Multi-domain learning performance across different ranks, with methods maintaining a comparable number of trainable parameters on the same x-axis. (b) Ablation study of hyperparameters $\beta$ and $d$ in common-sense multi-task learning. (c) Comparison of training latency for various PEFT methods with FastMoE.}
\label{fig: training Latency}
\vspace{-10pt}
\end{figure*}

To gain deeper insight into the contribution of each components in MALoRA, the result of the ablation study are presented in Table \ref{tag: ablation study}. 
Specifically, we compare MALoRA’s performance with various modifications to assess the impact of asymmetry, shared subspaces, and rank configurations.
For MoLoRA, initializing matrices $A_t$ in the same layer with a fixed random matrix shows a comparable performance against MoLoRA, suggesting that initializing experts similarly does not significantly damage multi-task learning performance. 
Even when experts in MALoRA are projected into an identical low-rank space $\text{span}(S_A)$, the model outperforms MoLoRA while reducing trainable parameters by 30\%, confirming the compressibility of MoLoRA.\looseness-1

For MALoRA variants, \textit{w fixed $S_A$}, which freezes $S_A$ and expands $\overline{r}$ to 16, leads to performance degradation. This implies that a randomly initialized $S_A$ which is not updated during training fails to capture meaningful task-specific features. The lack of adaptability prevents the model from aligning the subspace with task distributions.\looseness-1

\textit{w/o Asymmetry}, which use a symmetric structure of MALoRA with LoRA rank $r=8$ and $d=64$, shows that increasing the LoRA rank is more effective for improving learning capacity than expanding $d$. Even with the same number of trainable parameters, the asymmetric structure in MALoRA extends the generalization boundary.
In contrast, \textit{w/o shared $S_A$} that replaces the shared subspace $S_A$ with distinct down-projection matrices for each expert limits the rank of each expert’s matrix to $\frac{d}{N}$. This reduced rank constrains the generalization boundary and leads to degraded performance.\looseness-1

Freezing $P_t$ without fine-tuning leads to a notable decline in performance, as it prevents the model from fully leveraging the low-rank space $S_A$, thereby limiting its ability to adapt to task-specific features.
When MALoRA’s structure is applied to $B_t$ instead of $A_t$ in \textit{decomposing $B_t$}, performance decreases by 1.5\% on average. This compression of $B_t$ reduces the model’s expressive capacity, as $B_t$ has lower inter-expert similarity than $A_t$, highlighting the need for more adaptable representations in $B_t$ to handle task-specific variations effectively.\looseness-1

\paragraph{Rank Robustness} We investigates the effectiveness of MALoRA across various ranks $\overline{r}$. As shown in Figure \ref{fig: training Latency}(a), MALoRA consistently outperforms MoLoRA across all rank configurations, achieving significant improvements of up to 1.16\% at rank 4(6) (The values in parentheses represent the rank of MALoRA) and maintaining a clear advantage throughout. 
Most notably, MALoRA at rank $\overline{r}=12$ matches the performance of MoLoRA at a higher rank of $r=16$, while reducing trainable parameters by 48\%. This result underscores MALoRA’s ability to maintain high performance with fewer resources. 
The diminishing returns in MoLoRA as ranks increase indicate that simply increasing rank without efficient adaptation leads to overfitting at higher ranks. In contrast, MALoRA more effectively scales its capacity as rank increases. This is likely because its asymmetric structure focuses on fine-tuning the projection coefficients $P_t$ for task-specific adaptation, while maintaining the stability of the shared subspace $S_A$, helping to prevent overfitting in high-rank configurations.\looseness-1

\vspace{-5pt}
\paragraph{Hyperparameter Robustness} 
We conducted experiments to evaluate MALoRA’s sensitivity to hyperparameters $\beta$ and $d$, with changes to $d$ also adjusting $\overline{r}$ to maintain consistent trainable parameters. As shown in Figure 4(b), extreme values of $d$—too large ($d=64, \overline{r}=8$) or too small ($d=16, \overline{r}=14$)—lead to suboptimal results, emphasizing the need for a balanced allocation between the shared subspace and expert-specific components. The best performance occurs at $d=48, \overline{r}=10$, achieving a balance between shared and task-specific representations. MALoRA performs optimally when $\beta=1.25$, striking the ideal balance between refining task-specific features in $P_t$ and maintaining the stability of $S_A$. When $\beta > 1$, the model places more focus on task-specific learning, enhancing performance in scenarios requiring stronger task differentiation.\looseness-1

\paragraph{Latency} 
As shown in Figure \ref{fig: training Latency}(c), MALoRA reduces latency by 1.2x compared to MoLoRA during a single training step. By projecting inputs into the low-rank space $\text{span}(S_A)$, MALoRA decreases dimensionality in each MoE layer, reducing latency from gradient computations, data routing, and result collection. This results in significantly faster backward propagation due to more efficient gradient calculations in the compressed low-rank space. Although forward computations still have some complexity, the reduction in communication time compensates, leading to an overall time decrease per step. When integrated with FastMoE~\cite{he2021fastmoe}, both LoRA and MALoRA complete a training step in 4.3 seconds, while MoLoRA and DoRA take 4.65 and 4.97 seconds, respectively. The modular structure introduces a slight increase in optimization time, but the overall latency reduction achieved by MALoRA remains substantial. Ultimately, MALoRA’s efficient use of low-rank projections and modular design significantly reduces forward and backward propagation times, improving training speed without sacrificing performance.
Details of latency are provided in Appendix \ref{sec:Training Latency Comparison}.\looseness-1

\section{Related Work}
\subsection{Redundancy of PEFT Methods}
PEFT methods such as Prefix-Tuning \cite{li2021prefix}, Prompt-Tuning \cite{lester2021power}, and LoRA \cite{hu2021lora} are widely adopted to reduce the computational cost of fine-tuning LLMs. Recent research has highlighted two main areas for improvement in LoRA: structural inequity and redundancy. To address inequity, methods like LoRA+ \cite{hayou2024lora+}, Asymmetry LoRA \cite{zhu2024asymmetry}, and LoRA-FA \cite{zhang2023lora} adopts unequal fine-tuning strategy of LoRA modules. On the redundancy front, VeRA \cite{kopiczko2023vera} and VB-LoRA \cite{li2024vb} reduce the number of trainable parameters by sharing parameters across LoRA. MoLA \cite{gao2024higher} explores the optimal distribution of LoRA experts in MoLoRA, suggesting that higher layers benefit from a greater number of experts.\looseness-1

\subsection{Mixture-of-Experts}
MoE~\cite{jacobs1991adaptive} distributes data to sub-modules for processing via a routing mechanism, activating the neurons of the feed-forward sub-layers through sparse parameters. SpraseMoE  \cite{shazeer2017outrageously} and GShard \cite{lepikhin2020gshard} substantially expands the learning capacity of Transformers by incorporating numerous FFN experts within MoE framework.
LLMs based on MoE architectures, such as Mixtral 8x7B \cite{jiang2024mixtral}, have demonstrated remarkable performance. Hybrid models that integrate LoRA as experts aim to improve model capacity and better fit complex data during fine-tuning, paralleling the approach of MoE models. Relevant studies \cite{zadouri2023pushing, wu2024mixture} explore the potential of using LoRA as parameter-efficient experts across NLP and Vision application scenarios. MoRAL \cite{yang2024moral} leverages MoLoRA to enhance lifelong learning. Approaches like MoELoRA and MixLoRA \cite{luo2024moelora,li2024mixlora} employ MoLoRA to mitigate catastrophic forgetting in multi-task learning. The effectiveness of MoLoRA has been broadly validated, solidifying it as a key trend in PEFT research.\looseness-1

\section{Conclusion}

MALoRA introduces a novel PEFT approach for multi-task learning by leveraging a MoE structure with asymmetric low-rank adaptation. Through a shared down-projection space and expanded up-projection ranks, MALoRA optimizes parameter efficiency while enhancing generalization across tasks. Our experiments demonstrate that MALoRA consistently outperforms existing methods like LoRA and MoLoRA, particularly excelling in complex inter-domain and intra-domain scenarios. By reducing trainable parameters by up to 48\% and achieving a 1.2x increase in training speed, MALoRA addresses key challenges such as training imbalances and overfitting phenomenons observed in previous methods, offering a robust, scalable solution for diverse applications in multi-task learning.\looseness-1

\section{Limitation}
The proposed MALoRA method outperforms previous PEFT baselines while effectively reducing both training parameters and latency. However, to ensure that the enhancement of multi-task learning capability remains independent of the gating network, our approach maintains strict consistency in routing strategy and hyperparameter settings of MoLoRA. The optimization of MALoRA is network-agnostic, suggesting that customized routing strategies could further improve its performance and it is not involved in our research. We leave this exploration for future work. Moreover, multi-task fine-tuning may triggers data security and privacy issues in certain application scenarios. Further researches and developments in LLM safety are needed.\looseness-1
\bibliography{custom}

\begin{thebibliography}{47}
\providecommand{\natexlab}[1]{#1}

\bibitem[{Bisk et~al.(2020)Bisk, Zellers, Bras, Gao, and Choi}]{Bisk2020}
Yonatan Bisk, Rowan Zellers, Ronan~Le Bras, Jianfeng Gao, and Yejin Choi. 2020.
\newblock Piqa: Reasoning about physical commonsense in natural language.
\newblock In \emph{Thirty-Fourth AAAI Conference on Artificial Intelligence}.

\bibitem[{Chen et~al.(2021)Chen, Tworek, Jun, Yuan, de~Oliveira~Pinto, Kaplan, Edwards, Burda, Joseph, Brockman, Ray, Puri, Krueger, Petrov, Khlaaf, Sastry, Mishkin, Chan, Gray, Ryder, Pavlov, Power, Kaiser, Bavarian, Winter, Tillet, Such, Cummings, Plappert, Chantzis, Barnes, Herbert-Voss, Guss, Nichol, Paino, Tezak, Tang, Babuschkin, Balaji, Jain, Saunders, Hesse, Carr, Leike, Achiam, Misra, Morikawa, Radford, Knight, Brundage, Murati, Mayer, Welinder, McGrew, Amodei, McCandlish, Sutskever, and Zaremba}]{chen2021evaluating}
Mark Chen, Jerry Tworek, Heewoo Jun, Qiming Yuan, Henrique~Ponde de~Oliveira~Pinto, Jared Kaplan, Harri Edwards, Yuri Burda, Nicholas Joseph, Greg Brockman, Alex Ray, Raul Puri, Gretchen Krueger, Michael Petrov, Heidy Khlaaf, Girish Sastry, Pamela Mishkin, Brooke Chan, Scott Gray, Nick Ryder, Mikhail Pavlov, Alethea Power, Lukasz Kaiser, Mohammad Bavarian, Clemens Winter, Philippe Tillet, Felipe~Petroski Such, Dave Cummings, Matthias Plappert, Fotios Chantzis, Elizabeth Barnes, Ariel Herbert-Voss, William~Hebgen Guss, Alex Nichol, Alex Paino, Nikolas Tezak, Jie Tang, Igor Babuschkin, Suchir Balaji, Shantanu Jain, William Saunders, Christopher Hesse, Andrew~N. Carr, Jan Leike, Josh Achiam, Vedant Misra, Evan Morikawa, Alec Radford, Matthew Knight, Miles Brundage, Mira Murati, Katie Mayer, Peter Welinder, Bob McGrew, Dario Amodei, Sam McCandlish, Ilya Sutskever, and Wojciech Zaremba. 2021.
\newblock \href {https://arxiv.org/abs/2107.03374} {Evaluating large language models trained on code}.
\newblock \emph{Preprint}, arXiv:2107.03374.

\bibitem[{Chung et~al.(2024)Chung, Hou, Longpre, Zoph, Tay, Fedus, Li, Wang, Dehghani, Brahma et~al.}]{chung2024scaling}
Hyung~Won Chung, Le~Hou, Shayne Longpre, Barret Zoph, Yi~Tay, William Fedus, Yunxuan Li, Xuezhi Wang, Mostafa Dehghani, Siddhartha Brahma, et~al. 2024.
\newblock Scaling instruction-finetuned language models.
\newblock \emph{Journal of Machine Learning Research}, 25(70):1--53.

\bibitem[{Clark et~al.(2019)Clark, Lee, Chang, Kwiatkowski, Collins, and Toutanova}]{clark2019boolq}
Christopher Clark, Kenton Lee, Ming-Wei Chang, Tom Kwiatkowski, Michael Collins, and Kristina Toutanova. 2019.
\newblock Boolq: Exploring the surprising difficulty of natural yes/no questions.
\newblock In \emph{NAACL}.

\bibitem[{Clark et~al.(2018)Clark, Cowhey, Etzioni, Khot, Sabharwal, Schoenick, and Tafjord}]{allenai:arc}
Peter Clark, Isaac Cowhey, Oren Etzioni, Tushar Khot, Ashish Sabharwal, Carissa Schoenick, and Oyvind Tafjord. 2018.
\newblock Think you have solved question answering? try arc, the ai2 reasoning challenge.
\newblock \emph{arXiv:1803.05457v1}.

\bibitem[{Cobbe et~al.(2021)Cobbe, Kosaraju, Bavarian, Chen, Jun, Kaiser, Plappert, Tworek, Hilton, Nakano, Hesse, and Schulman}]{cobbe2021gsm8k}
Karl Cobbe, Vineet Kosaraju, Mohammad Bavarian, Mark Chen, Heewoo Jun, Lukasz Kaiser, Matthias Plappert, Jerry Tworek, Jacob Hilton, Reiichiro Nakano, Christopher Hesse, and John Schulman. 2021.
\newblock Training verifiers to solve math word problems.
\newblock \emph{arXiv preprint arXiv:2110.14168}.

\bibitem[{Dou et~al.(2024)Dou, Zhou, Liu, Gao, Shen, Xiong, Zhou, Wang, Xi, Fan et~al.}]{dou2024loramoe}
Shihan Dou, Enyu Zhou, Yan Liu, Songyang Gao, Wei Shen, Limao Xiong, Yuhao Zhou, Xiao Wang, Zhiheng Xi, Xiaoran Fan, et~al. 2024.
\newblock Loramoe: Alleviating world knowledge forgetting in large language models via moe-style plugin.
\newblock In \emph{Proceedings of the 62nd Annual Meeting of the Association for Computational Linguistics (Volume 1: Long Papers)}, pages 1932--1945.

\bibitem[{Du{\v{s}}ek et~al.(2020)Du{\v{s}}ek, Novikova, and Rieser}]{dusek.etal2020:csl}
Ond\v{r}ej Du{\v{s}}ek, Jekaterina Novikova, and Verena Rieser. 2020.
\newblock \href {https://doi.org/10.1016/j.csl.2019.06.009} {Evaluating the {{State}}-of-the-{{Art}} of {{End}}-to-{{End Natural Language Generation}}: {{The E2E NLG Challenge}}}.
\newblock \emph{Computer Speech \& Language}, 59:123--156.

\bibitem[{Gao et~al.(2024)Gao, Chen, Rao, Sun, Liu, Peng, Zhang, Guo, Yang, and Subrahmanian}]{gao2024higher}
Chongyang Gao, Kezhen Chen, Jinmeng Rao, Baochen Sun, Ruibo Liu, Daiyi Peng, Yawen Zhang, Xiaoyuan Guo, Jie Yang, and VS~Subrahmanian. 2024.
\newblock Higher layers need more lora experts.
\newblock \emph{arXiv preprint arXiv:2402.08562}.

\bibitem[{{Gaurang Bharti}(2024)}]{gaurang_bharti_2024}
{Gaurang Bharti}. 2024.
\newblock \href {https://doi.org/10.57967/hf/2557} {finance-alpaca (revision 51d16b6)}.

\bibitem[{Hayou et~al.(2024)Hayou, Ghosh, and Yu}]{hayou2024lora+}
Soufiane Hayou, Nikhil Ghosh, and Bin Yu. 2024.
\newblock Lora+: Efficient low rank adaptation of large models.
\newblock \emph{arXiv preprint arXiv:2402.12354}.

\bibitem[{He et~al.(2024)He, Luo, Hu, Zhao, Zhou, Wu, Zhang, Han, Liu, and Sun}]{he2024ultraeval}
Chaoqun He, Renjie Luo, Shengding Hu, Yuanqian Zhao, Jie Zhou, Hanghao Wu, Jiajie Zhang, Xu~Han, Zhiyuan Liu, and Maosong Sun. 2024.
\newblock \href {https://arxiv.org/abs/2404.07584} {Ultraeval: A lightweight platform for flexible and comprehensive evaluation for llms}.
\newblock \emph{Preprint}, arXiv:2404.07584.

\bibitem[{He et~al.(2021)He, Qiu, Zeng, Yang, Zhai, and Tang}]{he2021fastmoe}
Jiaao He, Jiezhong Qiu, Aohan Zeng, Zhilin Yang, Jidong Zhai, and Jie Tang. 2021.
\newblock Fastmoe: A fast mixture-of-expert training system.
\newblock \emph{arXiv preprint arXiv:2103.13262}.

\bibitem[{He et~al.(2015)He, Zhang, Ren, and Sun}]{he2015delving}
Kaiming He, Xiangyu Zhang, Shaoqing Ren, and Jian Sun. 2015.
\newblock Delving deep into rectifiers: Surpassing human-level performance on imagenet classification.
\newblock In \emph{Proceedings of the IEEE international conference on computer vision}, pages 1026--1034.

\bibitem[{Houlsby et~al.(2019)Houlsby, Giurgiu, Jastrzebski, Morrone, De~Laroussilhe, Gesmundo, Attariyan, and Gelly}]{houlsby2019parameter}
Neil Houlsby, Andrei Giurgiu, Stanislaw Jastrzebski, Bruna Morrone, Quentin De~Laroussilhe, Andrea Gesmundo, Mona Attariyan, and Sylvain Gelly. 2019.
\newblock Parameter-efficient transfer learning for nlp.
\newblock In \emph{International conference on machine learning}, pages 2790--2799. PMLR.

\bibitem[{Hu et~al.(2021)Hu, Shen, Wallis, Allen-Zhu, Li, Wang, Wang, and Chen}]{hu2021lora}
Edward~J Hu, Yelong Shen, Phillip Wallis, Zeyuan Allen-Zhu, Yuanzhi Li, Shean Wang, Lu~Wang, and Weizhu Chen. 2021.
\newblock Lora: Low-rank adaptation of large language models.
\newblock \emph{arXiv preprint arXiv:2106.09685}.

\bibitem[{Jacobs et~al.(1991)Jacobs, Jordan, Nowlan, and Hinton}]{jacobs1991adaptive}
Robert~A Jacobs, Michael~I Jordan, Steven~J Nowlan, and Geoffrey~E Hinton. 1991.
\newblock Adaptive mixtures of local experts.
\newblock \emph{Neural computation}, 3(1):79--87.

\bibitem[{Jiang et~al.(2023)Jiang, Sablayrolles, Mensch, Bamford, Chaplot, Casas, Bressand, Lengyel, Lample, Saulnier et~al.}]{jiang2023mistral}
Albert~Q Jiang, Alexandre Sablayrolles, Arthur Mensch, Chris Bamford, Devendra~Singh Chaplot, Diego de~las Casas, Florian Bressand, Gianna Lengyel, Guillaume Lample, Lucile Saulnier, et~al. 2023.
\newblock Mistral 7b.
\newblock \emph{arXiv preprint arXiv:2310.06825}.

\bibitem[{Jiang et~al.(2024)Jiang, Sablayrolles, Roux, Mensch, Savary, Bamford, Chaplot, Casas, Hanna, Bressand et~al.}]{jiang2024mixtral}
Albert~Q Jiang, Alexandre Sablayrolles, Antoine Roux, Arthur Mensch, Blanche Savary, Chris Bamford, Devendra~Singh Chaplot, Diego de~las Casas, Emma~Bou Hanna, Florian Bressand, et~al. 2024.
\newblock Mixtral of experts.
\newblock \emph{arXiv preprint arXiv:2401.04088}.

\bibitem[{Kopiczko et~al.(2023)Kopiczko, Blankevoort, and Asano}]{kopiczko2023vera}
Dawid~Jan Kopiczko, Tijmen Blankevoort, and Yuki~Markus Asano. 2023.
\newblock Vera: Vector-based random matrix adaptation.
\newblock \emph{arXiv preprint arXiv:2310.11454}.

\bibitem[{Le~Scao et~al.(2023)Le~Scao, Fan, Akiki, Pavlick, Ili{\'c}, Hesslow, Castagn{\'e}, Luccioni, Yvon, Gall{\'e} et~al.}]{le2023bloom}
Teven Le~Scao, Angela Fan, Christopher Akiki, Ellie Pavlick, Suzana Ili{\'c}, Daniel Hesslow, Roman Castagn{\'e}, Alexandra~Sasha Luccioni, Fran{\c{c}}ois Yvon, Matthias Gall{\'e}, et~al. 2023.
\newblock Bloom: A 176b-parameter open-access multilingual language model.

\bibitem[{Lepikhin et~al.(2020)Lepikhin, Lee, Xu, Chen, Firat, Huang, Krikun, Shazeer, and Chen}]{lepikhin2020gshard}
Dmitry Lepikhin, HyoukJoong Lee, Yuanzhong Xu, Dehao Chen, Orhan Firat, Yanping Huang, Maxim Krikun, Noam Shazeer, and Zhifeng Chen. 2020.
\newblock Gshard: Scaling giant models with conditional computation and automatic sharding.
\newblock \emph{arXiv preprint arXiv:2006.16668}.

\bibitem[{Lester et~al.(2021)Lester, Al-Rfou, and Constant}]{lester2021power}
Brian Lester, Rami Al-Rfou, and Noah Constant. 2021.
\newblock The power of scale for parameter-efficient prompt tuning.
\newblock \emph{arXiv preprint arXiv:2104.08691}.

\bibitem[{Li et~al.(2024{\natexlab{a}})Li, Ma, Wang, Cheng, Duan, Zuo, Yang, and Tang}]{li2024mixlora}
Dengchun Li, Yingzi Ma, Naizheng Wang, Zhiyuan Cheng, Lei Duan, Jie Zuo, Cal Yang, and Mingjie Tang. 2024{\natexlab{a}}.
\newblock Mixlora: Enhancing large language models fine-tuning with lora based mixture of experts.
\newblock \emph{arXiv preprint arXiv:2404.15159}.

\bibitem[{Li and Liang(2021)}]{li2021prefix}
Xiang~Lisa Li and Percy Liang. 2021.
\newblock Prefix-tuning: Optimizing continuous prompts for generation.
\newblock \emph{arXiv preprint arXiv:2101.00190}.

\bibitem[{Li et~al.(2024{\natexlab{b}})Li, Han, and Ji}]{li2024vb}
Yang Li, Shaobo Han, and Shihao Ji. 2024{\natexlab{b}}.
\newblock Vb-lora: Extreme parameter efficient fine-tuning with vector banks.
\newblock \emph{arXiv preprint arXiv:2405.15179}.

\bibitem[{Liu et~al.(2024{\natexlab{a}})Liu, Wu, Zhao, Zhu, Xu, Tian, and Zheng}]{liu2024moe}
Qidong Liu, Xian Wu, Xiangyu Zhao, Yuanshao Zhu, Derong Xu, Feng Tian, and Yefeng Zheng. 2024{\natexlab{a}}.
\newblock When moe meets llms: Parameter efficient fine-tuning for multi-task medical applications.
\newblock In \emph{Proceedings of the 47th International ACM SIGIR Conference on Research and Development in Information Retrieval}, pages 1104--1114.

\bibitem[{Liu et~al.(2024{\natexlab{b}})Liu, Wang, Yin, Molchanov, Wang, Cheng, and Chen}]{liu2024dora}
Shih-Yang Liu, Chien-Yi Wang, Hongxu Yin, Pavlo Molchanov, Yu-Chiang~Frank Wang, Kwang-Ting Cheng, and Min-Hung Chen. 2024{\natexlab{b}}.
\newblock Dora: Weight-decomposed low-rank adaptation.
\newblock \emph{arXiv preprint arXiv:2402.09353}.

\bibitem[{Liu and Luo(2024)}]{liu2024adamole}
Zefang Liu and Jiahua Luo. 2024.
\newblock Adamole: Fine-tuning large language models with adaptive mixture of low-rank adaptation experts.
\newblock \emph{arXiv preprint arXiv:2405.00361}.

\bibitem[{Luo et~al.(2024)Luo, Lei, Lei, Liu, He, Zhao, and Liu}]{luo2024moelora}
Tongxu Luo, Jiahe Lei, Fangyu Lei, Weihao Liu, Shizhu He, Jun Zhao, and Kang Liu. 2024.
\newblock Moelora: Contrastive learning guided mixture of experts on parameter-efficient fine-tuning for large language models.
\newblock \emph{arXiv preprint arXiv:2402.12851}.

\bibitem[{Malo et~al.(2014)Malo, Sinha, Korhonen, Wallenius, and Takala}]{Malo2014GoodDO}
P.~Malo, A.~Sinha, P.~Korhonen, J.~Wallenius, and P.~Takala. 2014.
\newblock Good debt or bad debt: Detecting semantic orientations in economic texts.
\newblock \emph{Journal of the Association for Information Science and Technology}, 65.

\bibitem[{Mihaylov et~al.(2018)Mihaylov, Clark, Khot, and Sabharwal}]{OpenBookQA2018}
Todor Mihaylov, Peter Clark, Tushar Khot, and Ashish Sabharwal. 2018.
\newblock Can a suit of armor conduct electricity? a new dataset for open book question answering.
\newblock In \emph{EMNLP}.

\bibitem[{Pal et~al.(2022)Pal, Umapathi, and Sankarasubbu}]{pmlr-v174-pal22a}
Ankit Pal, Logesh~Kumar Umapathi, and Malaikannan Sankarasubbu. 2022.
\newblock \href {https://proceedings.mlr.press/v174/pal22a.html} {Medmcqa: A large-scale multi-subject multi-choice dataset for medical domain question answering}.
\newblock In \emph{Proceedings of the Conference on Health, Inference, and Learning}, volume 174 of \emph{Proceedings of Machine Learning Research}, pages 248--260. PMLR.

\bibitem[{Peng et~al.(2023)Peng, Li, He, Galley, and Gao}]{peng2023instruction}
Baolin Peng, Chunyuan Li, Pengcheng He, Michel Galley, and Jianfeng Gao. 2023.
\newblock Instruction tuning with gpt-4.
\newblock \emph{arXiv preprint arXiv:2304.03277}.

\bibitem[{Ramsay et~al.(1984)Ramsay, ten Berge, and Styan}]{ramsay1984matrix}
James~O Ramsay, Jos ten Berge, and George~PH Styan. 1984.
\newblock Matrix correlation.
\newblock \emph{Psychometrika}, 49(3):403--423.

\bibitem[{Shazeer et~al.(2017)Shazeer, Mirhoseini, Maziarz, Davis, Le, Hinton, and Dean}]{shazeer2017outrageously}
Noam Shazeer, Azalia Mirhoseini, Krzysztof Maziarz, Andy Davis, Quoc Le, Geoffrey Hinton, and Jeff Dean. 2017.
\newblock Outrageously large neural networks: The sparsely-gated mixture-of-experts layer.
\newblock \emph{arXiv preprint arXiv:1701.06538}.

\bibitem[{Shi et~al.(2022)Shi, Suzgun, Freitag, Wang, Srivats, Vosoughi, Chung, Tay, Ruder, Zhou, Das, and Wei}]{shi2022language}
Freda Shi, Mirac Suzgun, Markus Freitag, Xuezhi Wang, Suraj Srivats, Soroush Vosoughi, Hyung~Won Chung, Yi~Tay, Sebastian Ruder, Denny Zhou, Dipanjan Das, and Jason Wei. 2022.
\newblock \href {https://arxiv.org/abs/2210.03057} {Language models are multilingual chain-of-thought reasoners}.
\newblock \emph{Preprint}, arXiv:2210.03057.

\bibitem[{Touvron et~al.(2023{\natexlab{a}})Touvron, Lavril, Izacard, Martinet, Lachaux, Lacroix, Rozi{\`e}re, Goyal, Hambro, Azhar et~al.}]{touvron2023llama2}
Hugo Touvron, Thibaut Lavril, Gautier Izacard, Xavier Martinet, Marie-Anne Lachaux, Timoth{\'e}e Lacroix, Baptiste Rozi{\`e}re, Naman Goyal, Eric Hambro, Faisal Azhar, et~al. 2023{\natexlab{a}}.
\newblock Llama: Open and efficient foundation language models.
\newblock \emph{arXiv preprint arXiv:2302.13971}.

\bibitem[{Touvron et~al.(2023{\natexlab{b}})Touvron, Martin, Stone, Albert, Almahairi, Babaei, Bashlykov, Batra, Bhargava, Bhosale et~al.}]{touvron2023llama}
Hugo Touvron, Louis Martin, Kevin Stone, Peter Albert, Amjad Almahairi, Yasmine Babaei, Nikolay Bashlykov, Soumya Batra, Prajjwal Bhargava, Shruti Bhosale, et~al. 2023{\natexlab{b}}.
\newblock Llama 2: Open foundation and fine-tuned chat models.
\newblock \emph{arXiv preprint arXiv:2307.09288}.

\bibitem[{Wei et~al.(2021)Wei, Bosma, Zhao, Guu, Yu, Lester, Du, Dai, and Le}]{wei2021finetuned}
Jason Wei, Maarten Bosma, Vincent~Y Zhao, Kelvin Guu, Adams~Wei Yu, Brian Lester, Nan Du, Andrew~M Dai, and Quoc~V Le. 2021.
\newblock Finetuned language models are zero-shot learners.
\newblock \emph{arXiv preprint arXiv:2109.01652}.

\bibitem[{Wei et~al.(2023)Wei, Wang, Liu, Ding, and Zhang}]{wei2023magicoder}
Yuxiang Wei, Zhe Wang, Jiawei Liu, Yifeng Ding, and Lingming Zhang. 2023.
\newblock Magicoder: Source code is all you need.
\newblock \emph{arXiv preprint arXiv:2312.02120}.

\bibitem[{Wu et~al.(2024)Wu, Huang, and Wei}]{wu2024mixture}
Xun Wu, Shaohan Huang, and Furu Wei. 2024.
\newblock Mixture of lora experts.
\newblock \emph{arXiv preprint arXiv:2404.13628}.

\bibitem[{Yang et~al.(2024)Yang, Ali, Wang, Hu, and Wang}]{yang2024moral}
Shu Yang, Muhammad~Asif Ali, Cheng-Long Wang, Lijie Hu, and Di~Wang. 2024.
\newblock Moral: Moe augmented lora for llms' lifelong learning.
\newblock \emph{arXiv preprint arXiv:2402.11260}.

\bibitem[{Yu et~al.(2023)Yu, Jiang, Shi, Yu, Liu, Zhang, Kwok, Li, Weller, and Liu}]{yu2023metamath}
Longhui Yu, Weisen Jiang, Han Shi, Jincheng Yu, Zhengying Liu, Yu~Zhang, James~T Kwok, Zhenguo Li, Adrian Weller, and Weiyang Liu. 2023.
\newblock Metamath: Bootstrap your own mathematical questions for large language models.
\newblock \emph{arXiv preprint arXiv:2309.12284}.

\bibitem[{Zadouri et~al.(2023)Zadouri, {\"U}st{\"u}n, Ahmadian, Ermi{\c{s}}, Locatelli, and Hooker}]{zadouri2023pushing}
Ted Zadouri, Ahmet {\"U}st{\"u}n, Arash Ahmadian, Beyza Ermi{\c{s}}, Acyr Locatelli, and Sara Hooker. 2023.
\newblock Pushing mixture of experts to the limit: Extremely parameter efficient moe for instruction tuning.
\newblock \emph{arXiv preprint arXiv:2309.05444}.

\bibitem[{Zhang et~al.(2023)Zhang, Zhang, Shi, Chu, and Li}]{zhang2023lora}
Longteng Zhang, Lin Zhang, Shaohuai Shi, Xiaowen Chu, and Bo~Li. 2023.
\newblock Lora-fa: Memory-efficient low-rank adaptation for large language models fine-tuning.
\newblock \emph{arXiv preprint arXiv:2308.03303}.

\bibitem[{Zhu et~al.(2024)Zhu, Greenewald, Nadjahi, Borde, Gabrielsson, Choshen, Ghassemi, Yurochkin, and Solomon}]{zhu2024asymmetry}
Jiacheng Zhu, Kristjan Greenewald, Kimia Nadjahi, Haitz S{\'a}ez de~Oc{\'a}riz Borde, Rickard~Br{\"u}el Gabrielsson, Leshem Choshen, Marzyeh Ghassemi, Mikhail Yurochkin, and Justin Solomon. 2024.
\newblock Asymmetry in low-rank adapters of foundation models.
\newblock \emph{arXiv preprint arXiv:2402.16842}.

\end{thebibliography}

\appendix

\clearpage
\section{Appendix}
\label{sec:appendix}

\subsection{Datasets}
\label{sec:Datasets}
\subsubsection{Inter-Domain Tasks Learning}
\begin{table}[htb]
\centering
\renewcommand{\arraystretch}{1.4}
\resizebox{\columnwidth}{!}{%
\begin{tabular}{lcccc}
\specialrule{1pt}{0pt}{0pt}
\hline
\textbf{Task} & \textbf{Domain} &  \textbf{\#Train} & \textbf{\#Sample} & \textbf{Type} \\ \hline
MedMCQA \cite{pmlr-v174-pal22a} & Medicine & 182K & \multirow{6}{*}{30,000} & Question Answering \\
Magicoder \cite{wei2023magicoder} & Code & 110K & & Code Generation  \\
Finance Alpaca \cite{gaurang_bharti_2024} & Finance & 69K &  & Instruction Following \\
MetaMathQA \cite{yu2023metamath} & Mathematics & 395K & & Math Reasoning\\
Alpaca-GPT4 \cite{peng2023instruction}& General & 52K & & Instruction Following \\
E2E NLG \cite{dusek.etal2020:csl}& End-to-End & 34K & & Natural Language Generation \\
\hline
\specialrule{1pt}{0pt}{0pt}
\end{tabular}
}

\caption{Description of Inter-Domain Training Datasets. We sampled 30,000 training data from each dataset to form a mixture inter-domain multi-task training set.}
\end{table}

\begin{table}[htb]
\centering
\renewcommand{\arraystretch}{1.4}
\resizebox{\columnwidth}{!}{%
\begin{tabular}{lccc}
\specialrule{1pt}{0pt}{0pt}
\hline
\textbf{Task} & \textbf{Domain} &  \textbf{\#Test} & \textbf{Type} \\ \hline
MedMCQA & Medicine & 2,802\textsuperscript{†} & Question Answering \\
HumanEval \cite{chen2021evaluating} & Code & 164 & Code Generation  \\
PhraseBank \cite{Malo2014GoodDO} & Finance & 3,453 & Text Classification \\
GSM8K \cite{cobbe2021gsm8k} & Mathematics & 1,319 & Math Reasoning \\
ARC-C \cite{allenai:arc} & Commen-Sense & 1,170 &  Question Answering \\
ARC-E \cite{allenai:arc} & Commen-Sense & 2,380 &  Question Answering \\
E2E NLG & End-to-End & 4,693 & Natural Language Generation \\
\hline
\specialrule{1pt}{0pt}{0pt}
\end{tabular}
}

\caption{Description of Inter-Domain Testing Datasets. \textsuperscript{†} Due to the non-public nature of the labels for the MedMCQA test set, the development dataset is utilized for testing in UltraEval \cite{he2024ultraeval}, and entries with anomalies in multiple-choice selections or variations in the number of options have been cleaned.}
\end{table}

\subsubsection{Intra-Domain Tasks Learning}
\begin{table}[htb]
\centering
\renewcommand{\arraystretch}{1.4}
\resizebox{\columnwidth}{!}{%
\begin{tabular}{lccc}
\specialrule{1pt}{0pt}{0pt}
\hline
\textbf{Task} &  \textbf{\#Train} & \textbf{\#Test} & \textbf{Type} \\ \hline
ARC-C \cite{allenai:arc} & 1,120 & 1,170 & \multirow{5}{*}{Question Answering} \\
ARC-E \cite{allenai:arc} & 2,250 & 2,380 & \\
PIQA \cite{Bisk2020} & 16,100 & 1,840 & \\
OpenBookQA \cite{OpenBookQA2018} & 4,957 & 500 & \\
BoolQ \cite{clark2019boolq} & 9,427 & 3,270 & \\
\hline
\specialrule{1pt}{0pt}{0pt}
\end{tabular}
}
\caption{Description of Intra-Domain Training Datasets. Datasets focus on the scope of common-sense reasoning, following a common setting of existed works \cite{li2024mixlora, luo2024moelora}.}
\end{table}

\subsubsection{Quantities of Training Tokens}
\label{sec:Quantity of Training Tokens}
\begin{table}[htb]
\centering
\renewcommand{\arraystretch}{1.2}
\resizebox{\columnwidth}{!}{%
\begin{tabular}{lcccc}
\specialrule{1pt}{0pt}{0pt}
\hline
\textbf{Task} & \textbf{\#Input} & \textbf{\#Label} & \textbf{\#Total}  \\ \hline
MedMCQA & 469K & 157K & 626K  \\
Magicoder & 3,231K & \textbf{1,446K} & 4,677K   \\
Finance Alpaca & 839K & 206K & 1,045K  \\
MetaMathQA & 1,415K & 537K & 1,952K \\
Alpaca-GPT4 & 483K & 214K & 697K \\
E2E NLG & 202K & 461K & 663K \\ \hline
ARC-C & 18.5K & 1.8K & 20.3K  \\
ARC-E & 36.4K & 4.3K & 40.7K   \\
BoolQ & 378.7K & 18.7K & 397.4K  \\
OpenBookQA & 79.8K & 9.7K & 89.5K \\
PIQA & 318.1K & \textbf{31.9K} & 350K \\
\hline
\specialrule{1pt}{0pt}{0pt}
\end{tabular}
}
\caption{Description of Token Quantity in Each Dataset}

\label{tag: tokens quantity}
\end{table}

Table \ref{tag: tokens quantity} presents the token quantities distribution of training datasets. In the context of multi-domain learning, the Magicoder dataset contains a substantial 1,446K of label tokens for computing cross-entropy and performing backward propagation, significantly surpassing datasets in other domains. As shown in Table \ref{tag: cross domain result}, non-MoE PEFT methods achieve performance on the HumanEval benchmark that is comparable to single-task fine-tuning. However, on the Finance Alpaca dataset, which has fewer amount of label tokens, the learning efficacy of non-MoE methods is sub-optimal compared to MoE approaches. A similar trend is observed in common-sense multi-task learning, as indicated in Table \ref{tag: common sense result}. Due to the disparity in the magnitude of label tokens, LoRA demonstrates superior performance on the PIQA benchmark while exhibiting deficiencies on the ARC task. This highlights a clear correlation between the abundance of label tokens and LoRA's performance in the corresponding tasks. In summary, LoRA shows drawbacks related to learning imbalance and catastrophic forgetting.

\subsection{Hyperparameter Configuration}
\label{sec:Hyper-parameter Configuration}
\subsubsection{Inter-Domain Tasks Learning}
\begin{table}[H]
\centering
\renewcommand{\arraystretch}{1.4}
\resizebox{\columnwidth}{!}{%
\begin{tabular}{lcccc}
\specialrule{1pt}{0pt}{0pt}
\hline
\textbf{Hyperparameters} & \textbf{LoRA/DoRA} &  \textbf{AsyLoRA} & \textbf{MoLoRA} & \textbf{MALoRA} \\ \hline
Optimizer & \multicolumn{4}{c}{AdamW} \\
Learning Rate & \multicolumn{4}{c}{5e-4} \\
Epochs & \multicolumn{4}{c}{3} \\ 
Batch Size & \multicolumn{4}{c}{4} \\
Dropout & \multicolumn{4}{c}{0.05} \\ 
Weight Decay & \multicolumn{4}{c}{0.01} \\ 
Warm-up Ratio & \multicolumn{4}{c}{0.1} \\ \hline
Implement Layer & \multicolumn{4}{c}{Q,K,V,Up,Down,Gate} \\
LoRA Rank $r$ & 64 & 128 & 8 & 12 \\
LoRA Alpha $\alpha$ & 128 & 256 & 16 & 24 \\ \hline
Experts Number & \multicolumn{2}{c}{-} & \multicolumn{2}{c}{8} \\
Top-K & \multicolumn{2}{c}{-} & \multicolumn{2}{c}{2} \\
Balance Loss Factor & \multicolumn{2}{c}{-} & \multicolumn{2}{c}{0.01} \\  \hline
Projection Rank $d$ & \multicolumn{3}{c}{-} & 32 \\ 
Weight Balance Scale $\beta$ & \multicolumn{3}{c}{-} & 1 \\
\hline
\specialrule{1pt}{0pt}{0pt}
\end{tabular}
}
\caption{Experimental Hyperparameter Configurations for Inter-Domain Learning with Various PEFT Methods. Experiments are conducted on four Nvidia A100 GPUs.}
\end{table}

\subsubsection{Intra-Domain Tasks Learning}
\label{sec:Commen-sense Tasks Learning}
\begin{table}[H]
\centering
\renewcommand{\arraystretch}{1.4}
\resizebox{\columnwidth}{!}{%
\begin{tabular}{lcccc}
\specialrule{1pt}{0pt}{0pt}
\hline
\textbf{Hyperparameters} & \textbf{LoRA/DoRA} &  \textbf{AsyLoRA} & \textbf{MoLoRA} & \textbf{MALoRA} \\ \hline
Optimizer & \multicolumn{4}{c}{AdamW} \\
Learning Rate & \multicolumn{4}{c}{4e-4} \\
Epochs & \multicolumn{4}{c}{2} \\ 
Batch Size & \multicolumn{4}{c}{2} \\
Dropout & \multicolumn{4}{c}{0.05} \\ 
Weight Decay & \multicolumn{4}{c}{0.01} \\ 
Warm-up Ratio & \multicolumn{4}{c}{0.1} \\ \hline
Implement Layer & \multicolumn{4}{c}{Q,K,V,Up,Down,Gate} \\
LoRA Rank $r$ & 64 & 128 & 8 & 12 \\
LoRA Alpha $\alpha$ & 128 & 256 & 16 & 24 \\ \hline
Experts Number & \multicolumn{2}{c}{-} & \multicolumn{2}{c}{8} \\
Top-K & \multicolumn{2}{c}{-} & \multicolumn{2}{c}{2} \\
Balance Loss Factor & \multicolumn{2}{c}{-} & \multicolumn{2}{c}{0.01} \\  \hline
Projection Rank $d$ & \multicolumn{3}{c}{-} & 32 \\ 
Weight Balance Scale $\beta$ & \multicolumn{3}{c}{-} & 1.25 \\
\hline
\specialrule{1pt}{0pt}{0pt}
\end{tabular}
}

\caption{Experimental Hyperparameter Configurations for Common-Sense Multi-Task Learning with Various PEFT Methods. Experiments are conducted on a single NVIDIA A100 GPU.}
\end{table}

\subsection{The Impact of Hyperparameter $\beta$}
\label{sec:Impact of beta}
Defining $\mathcal{L}$ as the loss function of fine-tuning, $W \in \mathbb{R}^{m\times n}$ as a linear matrix in the backbone model where the MALoRA modules are attached to. We have the partial derivatives of loss $\mathcal{L}$ for matrix $S_A \in \mathbb{R}^{d\times n}$ and $P_t \in \mathbb{R}^{r\times d}$:\looseness-1
\begin{align}\label{eq:der}
\Delta W_t &= \overline{B}_t P_t S_A \\
\frac{\partial \mathcal{L}}{\partial S_A} &= \frac{\partial W}{\partial S_A}\frac{\partial \mathcal{L}}{\partial W},\ \ \frac{\partial \mathcal{L}}{\partial P_t} = \frac{\partial \mathcal{L}}{\partial W}\frac{\partial W}{\partial P_t} \\
\frac{\partial \mathcal{L}}{\partial S_A} &= \sum_{t=1}^N G_t(I^T\otimes (P_t\overline{B}_t))_{dn\times mn}\frac{\partial \mathcal{L}}{\partial W}\\
\frac{\partial \mathcal{L}}{\partial P_t} &= G_t\frac{\partial \mathcal{L}}{\partial W}(\overline{B}_t^T\otimes S_A)_{mn\times rd}
\end{align}

Here, operator $\otimes$ stands for the Kronecker Product. Since the norm of $P_t$ is proportional to $1/\beta$, and the norm of $S_A$ is proportional to $\beta$, so does the norm of gradient:
\begin{align}\label{eq:der2}
|\frac{\partial \mathcal{L}}{\partial S_A}| & \propto \frac{1}{\beta},\ \ |\frac{\partial \mathcal{L}}{\partial P_t}| \propto \beta
\end{align}

Hence, an elevated value of $\beta$ leads to an augmentation in the magnitude of $|\nabla P_t|$ and a concomitant reduction in the magnitude of $|\nabla S_A|$, thereby enhancing the model's capacity to discern task-specific discrimination. In contrast, a diminished value of $\beta$ endows the model with a propensity to primarily capture commonalities across datasets.\looseness-1

\subsection{Insufficient Learning Ability of AsyLoRA}
\label{sec:Insufficient Learning Ability of AsyLoRA}
\begin{figure}[htb]
    \centering
    \includegraphics[width=0.7\linewidth]{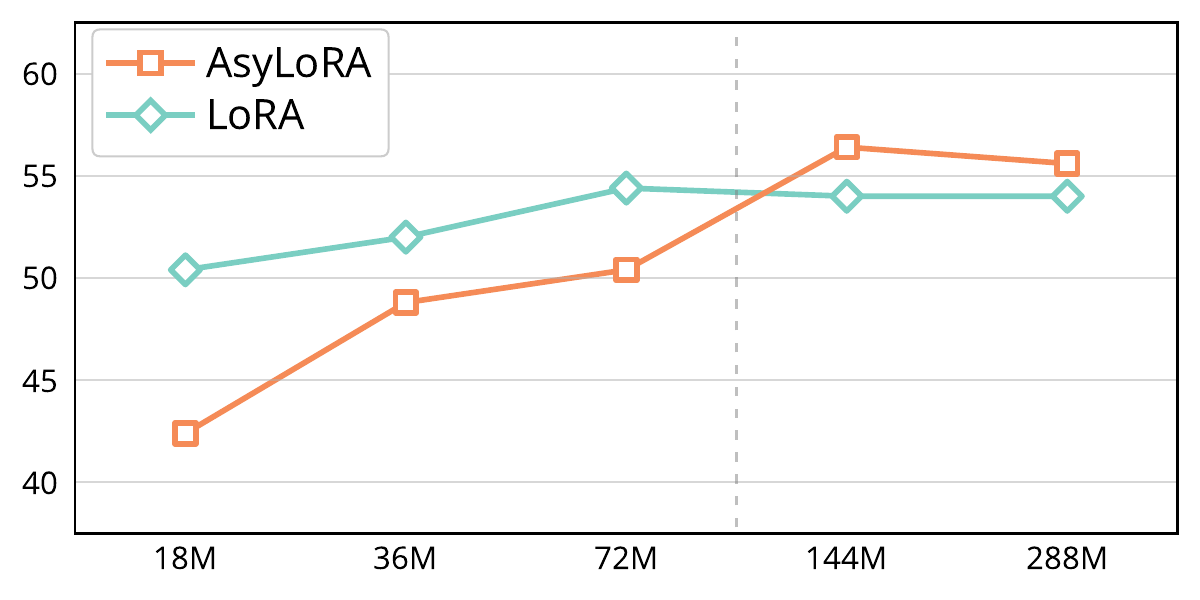}
    \caption{Performance Variations of LoRA and Asymmetry LoRA with Respect to Rank on the Math Reasoning Task MGSM \cite{shi2022language}.The ticks on the x-axis represent the number of trainable parameters in Asymmetry LoRA and LoRA.}
    \label{fig: Performance variants of LoRA and Asymmetry LoRA}
    \vspace{-10pt}
\end{figure}

In this section, we discuss the reasons for the poor performance of Mixture-of-Asymmetry-LoRA (MoAsyLoRA). In experts of MoAsyLoRA, the rank $r$ is doubled while the matrix $A$ remains frozen during fine-tuning. Consequently, the generalization boundary of these experts aligns with MoLoRA rather than expanding. To further investigate the mechanism behind performance degradation, we conduct an ablation study on the rank $r$ for both Asymmetry LoRA and LoRA methods in math reasoning tasks, as shown in Figure \ref{fig: Performance variants of LoRA and Asymmetry LoRA}. Under complex tasks like math reasoning, Asymmetry LoRA achieves better generalization only when the rank is sufficiently high. This can be attributed to the frozen down projection matrix, which randomly projects information dimensions in Asymmetry LoRA. When the model lacks the capability to fine-tune effectively, it compensates by increasing the dimension of the random projection, overriding the optimization of the down projection matrix in LoRA. However, experts in MoLoRA and MoAsyLoRA typically operate at low ranks, resulting in significant generalization drawbacks between Asymmetry LoRA and LoRA experts.

\subsection{Training Latency Comparison}
\label{sec:Training Latency Comparison}
\begin{table}[H]
\centering
\renewcommand{\arraystretch}{1.4}
\resizebox{\columnwidth}{!}{%
\begin{tabular}{lccccc}
\specialrule{1pt}{0pt}{0pt}
\hline
\multirow{2}{*}{\textbf{Method}} &  \multirow{2}{*}{\textbf{Forward}} & \multirow{2}{*}{\textbf{Backward}} & \multirow{2}{*}{\textbf{Optimize}} & \multicolumn{2}{c}{\textbf{Total}}\\ \cline{5-6}
  &    &   &   & Token(\(\mu s\)) & Step(\(s\))\\ \hline
LoRA & 4.246 & 9.229 & 0.188 & 833.9 & 13.663\\
DoRA & 4.973 & 11.514 & 0.240 & 1020.8 & 16.726\\
MoLoRA & 4.649 & 11.902 & 1.017 & 1072.3 & 17.568\\
MALoRA & 4.288 & 9.523 & 0.876 & 896.4 & 14.687\\
\hline
\specialrule{1pt}{0pt}{0pt}
\end{tabular}
}

\caption{Training Latency Components of PEFT Methods. A single step of training processes 16,384 input tokens. The single-step training latency of MALoRA accounts for only about 85\% of that of MoLoRA, achieving a 1.2x speed improvement. Metrics are evaluated using four NVIDIA A100 GPUs.}
\end{table}

\onecolumn
\subsection{Rank Robustness}
\label{sec:Rank Variation Ablation Study of MALoRA}
Table \ref{tag:rank ablation} shows the performance of MoLoRA and proposed MALoRA under proportional ranks. As the quantity of PEFT parameters increases, the dimension of $S_A$ expands, which in turn diminishes the marginal benefit of fine-tuning $S_A$, potentially culminating in over-fitting. Consequently, it is advisable to select larger $\beta$ to emphasize the acquisition of distinctive representations across diverse tasks. From this perspective, by scaling $\beta$ in proportion to the increase in rank $r$, we can effectively mitigate under-fitting and over-fitting phenomenas.

\setlength{\tabcolsep}{2mm}{ 
\setlength{\LTcapwidth}{\textwidth}   
\small
\centering
\renewcommand{\arraystretch}{1.35}
\begin{longtable}{lccccccccccc}
 \specialrule{2pt}{0pt}{0pt}
Method                       & $r$ & $d$ & $\beta$ & MedMCQA & HumanEval & GSM8K & PhraseBank & ARC-C & ARC-E & E2E & AVG. \\ \hline
\multirow{4}{*}{MoLoRA}    & 4 & \multirow{4}{*}{-} & \multirow{4}{*}{-}& 42.4 & 23.8 & 34.0 & 69.3 & 58.8 & 77.4 & 66.1 & 53.12       \\
    & 8 & & & 42.1 & 29.3 & 33.1 & 78.1 & 61.1 & 78.1 & 65.6 & 55.33  \\
    & 16 & & & 43.4 & 32.9 & 33.4 & 79.4 & 61.5 & 78.0 & 66.0 &  56.19          \\
    & 32 & & & 45.8 & 33.5 & 32.0 & 72.6 & 61.9 & 78.3 & 66.0 & 55.74      \\  \hline
\multirow{4}{*}{MALoRA}         & 6 & 16 & 0.3 & 43.2 & 29.9 & 32.0 & 74.0 & 58.9 & 76.7 & 65.5 & 54.28     \\
    & 12 & 32 & 1.0 & 42.3 & 32.9 & 34.1 & 79.8 & 60.0 & 78.5 & 66.3 & 56.27      \\
    & 24 & 64 & 3.5 & 42.0 & 30.5 & 36.1 & 80.6 & 62.4 & 78.6 & 66.1 & 56.61      \\
    & 48 & 128 & 5.0 & 43.9 & 34.8 & 35.2 & 76.7 & 62.2 & 78.1 & 65.9 & 56.69      \\
 \specialrule{2pt}{0pt}{0pt}
\caption{Performance of MoLoRA and Proposed MALoRA Under Proportional Ranks. In all configurations, MALoRA consistently outperformed the MoLoRA approach.}

 \label{tag:rank ablation}
\end{longtable}
}

\subsection{Hyperparameter Robustness}
\label{sec:Hyper-parameter Robustness of MALoRA}

\setlength{\tabcolsep}{5.4mm}{ 
\setlength{\LTcapwidth}{\textwidth}   
\small
\renewcommand{\arraystretch}{1.35}
\begin{center}
\begin{longtable}{ccccccccc}
 \specialrule{2pt}{0pt}{0pt}
$r$ & $d$   & $\beta$           & PIQA & OBQA & BoolQ  & ARC-C & ARC-E & AVG. \\ \hline \endhead
\multirow{4}{*}{14}   & \multirow{4}{*}{16} &  0.5 & 81.99 & 89.8 & 73.15 & 66.21 & 82.83 & 78.80 \\
    & & 1.0 & 81.28 & 87.4 & 79.69 & 65.74 & 81.31 & 79.08 \\ 
    & & 1.25 & 80.79 & 88.6 & 80.34 & 65.67 & 83.12 & 79.70 \\ 
    & & 2.0 & 79.98 & 87.8 & 78.44 & 66.89 & 81.65 & 78.95 \\ \hline
\multirow{4}{*}{12}   & \multirow{4}{*}{32} &  0.5 & 79.76 & 89.6 & 82.87 & 64.17 & 81.82 & 79.64 \\
    & & 1.0 & 80.96 & 89.0 & 81.38 & 66.55 & 82.24 & 80.03 \\ 
    & & 1.25 & 81.83 & 89.6 & 81.56 & 66.55 & 82.83 & 80.47 \\ 
    & & 2.0 & 81.66 & 88.4 & 78.01 & 65.74 & 82.37 & 79.24 \\ \hline
\multirow{4}{*}{10}   & \multirow{4}{*}{48} &  0.5 & 82.26 & 89.6 & 77.77 & 65.81 & 81.94 & 79.48 \\
    & & 1.0 & 82.32 & 89.0 & 76.54 & 68.12 & 82.62 & 79.72 \\ 
    & & 1.25 & 81.34 & 90.0 & 81.83 & 68.05 & 82.74 & 80.79 \\ 
    & & 2.0 & 81.56 & 88.4 & 82.17 & 67.78 & 82.58 & 80.50 \\ \hline
\multirow{4}{*}{8}   & \multirow{4}{*}{64} &  0.5 & 79.11 & 88.8 & 81.01 & 64.17 & 79.04 & 78.43 \\
    & & 1.0 & 79.87 & 87.6 & 83.33 & 63.90 & 80.77 & 79.09 \\ 
    & & 1.25 & 81.23 & 87.8 & 77.98 & 65.40 & 82.15 & 78.91 \\ 
    & & 2.0 & 80.14 & 87.8 & 81.83 & 64.79 & 80.30 & 78.97 \\ \hline
 \specialrule{2pt}{0pt}{0pt}
 
\caption{Details of Hyperparameter Robustness Experiment of MALoRA Evaluated on Intra-Domain Tasks.}
\end{longtable}
\end{center}
}

\end{document}